\documentclass[10pt,twocolumn,letterpaper]{article}

\usepackage{cvpr}              
\definecolor{cvprblue}{rgb}{0.21,0.49,0.74}
\usepackage[pagebackref,breaklinks,colorlinks,allcolors=cvprblue]{hyperref}

\def\paperID{19211} 
\def\confName{CVPR}
\def\confYear{2026}

\title{TextOVSR: Text-Guided Real-World Opera Video Super-Resolution}

\author{
Hua Chang\textsuperscript{1} ~
Xin Xu\textsuperscript{1,2,}\thanks{Corresponding authors} ~
Wei Liu\textsuperscript{1} ~
Jiayi Wu\textsuperscript{1} ~
Kui Jiang\textsuperscript{3,}\footnotemark[1] ~
Fei Ma\textsuperscript{4} ~
Qi Tian\textsuperscript{5} \\  
\textsuperscript{1}School of Computer Science and Technology, Wuhan University of Science and Technology, Wuhan, China \\
\textsuperscript{2}Hubei Province Key Laboratory of Intelligent Information Processing and Real-time Industrial System \\
\textsuperscript{3}Harbin Institute of Technology Zhengzhou Research Institute, Zhengzhou, China \\
\textsuperscript{4}Guangdong Laboratory of Artificial Intelligence and Digital Economy (SZ)\\
\textsuperscript{5}Huawei Technologies Ltd. \\
{\tt\small \{changhua, xuxin, liuwei, wuaddone\}@wust.edu.cn} \\ 
{\tt\small jiangkui@hit.edu.cn, mafei@gml.ac.cn, tian.qi1@huawei.com} 
}

\usepackage{amsmath}

\usepackage{indentfirst}
\usepackage{xcolor}
\usepackage{multirow}
\usepackage{booktabs}
\usepackage{adjustbox}
\usepackage{makecell}
\usepackage{amssymb}
\usepackage{utfsym}
\usepackage{xcolor}
\usepackage[table]{xcolor}
\usepackage{colortbl}
\usepackage{tabularray}
\usepackage{makecell}
\usepackage{graphicx}
\usepackage{enumitem}

\usepackage{subcaption}

\usepackage[accsupp]{axessibility}
\UseTblrLibrary{booktabs}

\begin{document}
\maketitle
\begin{abstract}

Many classic opera videos exhibit poor visual quality due to the limitations of early filming equipment and long-term degradation during storage. Although real-world video super-resolution (RWVSR) has achieved significant advances in recent years, directly applying existing methods to degraded opera videos remains challenging. 
The difficulties are twofold. First, accurately modeling real-world degradations is complex: simplistic combinations of classical degradation kernels fail to capture the authentic noise distribution, while methods that extract real noise patches from external datasets are prone to style mismatches that introduce visual artifacts. Second, current RWVSR methods, which rely solely on degraded image features, struggle to reconstruct realistic and detailed textures due to a lack of high-level semantic guidance. 
To address these issues, we propose a Text-guided Dual-Branch Opera Video Super-Resolution (TextOVSR) network, which introduces two types of textual prompts to guide the super-resolution process. Specifically, degradation-descriptive text, 
derived from the degradation process, is incorporated into the negative branch to constrain the solution space. 
Simultaneously, content-descriptive text is incorporated into 
a positive branch and our proposed Text-Enhanced Discriminator (TED) to provide semantic guidance for enhanced texture reconstruction. 
Furthermore, we design a Degradation-Robust Feature Fusion (DRF) module to facilitate cross-modal feature fusion while suppressing degradation interference. Experiments on our OperaLQ benchmark show that TextOVSR outperforms state-of-the-art methods both qualitatively and quantitatively. The code is available at \href{https://github.com/ChangHua0/TextOVSR}{https://github.com/ChangHua0/TextOVSR}. 


\end{abstract}
    
\section{Introduction}
\label{introduction}


Many classic opera videos exhibit poor visual quality due to limitations in early filming equipment and complex degradation processes during storage and transmission. Video super-resolution (VSR) has advanced rapidly in recent years \cite{chan2022basicvsr++,liang2024vrt}, with the primary goal of restoring high-resolution (HR) frames from their low-resolution (LR) counterparts, typically generated by known kernels (e.g., bicubic). However, low-quality videos in the real-world are not the product of simple downsampling; they are affected by a complex combination of unknown degradations, including sensor noise, compression artifacts, and transmission losses \cite{wang2021real}. Consequently, traditional VSR methods, trained under idealized degradation assumptions, suffer from a significant domain gap and demonstrate poor generalization when applied to real degraded videos \cite{chan2022investigating,song2024negvsr}.

\begin{figure*}[!t]
  \centering
   \includegraphics[width=1\linewidth]{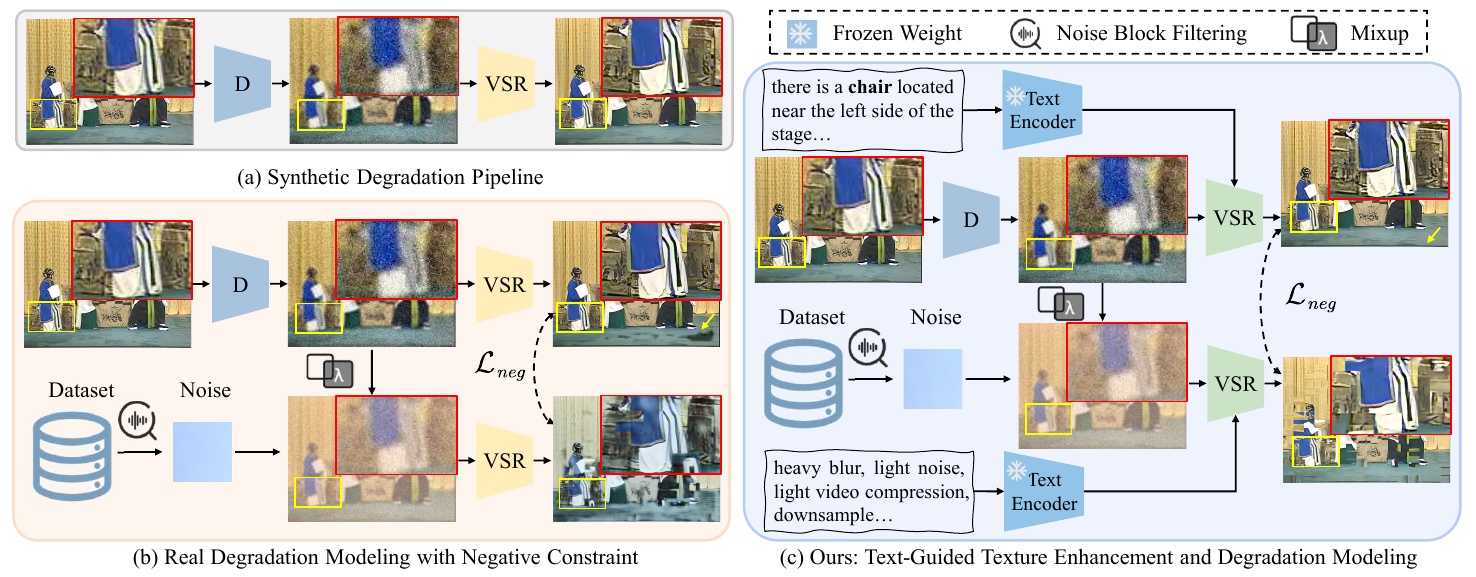}\vspace*{-2mm}
   \caption{\textbf{Frameworks for real-world video super-resolution.} (a) The classical synthetic degradation pipeline (D) simulates real degradations for the VSR model. (b) Real degradation modeling extracts authentic noise from external datasets and applies a negative constraint ($\mathcal L_{neg}$) to enhance robustness. (c) Our proposed TextOVSR introduces text-guided priors to enrich image features and model diverse degradations in the feature space. $\lambda$ controls the mixing ratio.}
   \label{fig1}\vspace*{-4mm}
\end{figure*}

Real-world video super-resolution (RWVSR) has been proposed to address this domain shift \cite{pan2021deep,bai2024self}. A central focus of this field is expanding the realism and diversity of the degradation space. For instance, Real-ESRGAN \cite{wang2021real} employs a high-order degradation pipeline that combines simple kernels to synthesize complex, low-quality images, as shown in \Cref{fig1}(a). A key limitation of this approach is its confinement to a limited, synthetic degradation space, which often results in out-of-distribution noise. More recently, NegVSR \cite{song2024negvsr} circumvented this by extracting real noise patches from external datasets to simulate authentic degradations (\Cref{fig1}(b)). However, this method is highly dependent on the stylistic similarity between the external and target datasets; a mismatch can introduce severe artifacts, as indicated by the yellow arrows in \Cref{fig1}. Furthermore, most existing RWVSR methods rely solely on degraded image features, which lack the high-level semantic information necessary for reconstructing realistic textures.

To achieve more effective degradation modeling and detail reconstruction, we propose TextOVSR, a dual-branch text-guided framework built upon NegVSR \cite{song2024negvsr} (\Cref{fig1}(c)). We synthesize degraded inputs using a high-order degradation pipeline \cite{wang2021real,chan2022investigating}. A key innovation is the generation of textual prompts to guide the super-resolution. We employ a binning method to qualitatively describe the severity of each degradation component, concatenating these into a comprehensive degradation-descriptive text. This text is incorporated into the negative branch; after encoding by a CLIP text encoder \cite{radford2021learning}, it guides the application of contrastive constraints to improve model robustness. To enhance texture reconstruction, the positive branch is guided by content-descriptive text, generated from pristine images using a multimodal large language model (MLLM) and fused with visual features via a novel Degradation-Robust Feature Fusion (DRF) module. The DRF module is designed to facilitate effective cross-modal fusion while suppressing interference from inherent degradations and style inconsistencies. Finally, we introduce a Text-Enhanced Discriminator (TED) that leverages high-level semantic cues from the content text to provide more accurate adversarial guidance, steering the generator toward photorealistic outputs. Extensive experiments on our self-constructed OperaLQ benchmark, comprising real-world degraded opera videos, demonstrate that TextOVSR achieves state-of-the-art performance in both qualitative and quantitative evaluations. In summary, our contributions are as follows:

\begin{itemize}
\item[$\bullet$]
We propose TextOVSR, a text-guided dual-branch network for opera video super-resolution, leveraging content- and degradation-descriptive texts to improve texture reconstruction and handle complex degradations.
\item[$\bullet$] 
We design a Degradation-Robust Feature Fusion (DRF) module that enables effective cross-modal fusion of visual and textual features while mitigating degradation-induced contamination.
\item[$\bullet$] 
We introduce a Text-Enhanced Discriminator (TED) that leverages text semantics to improve discrimination and guide more realistic outputs.
\item[$\bullet$] We construct and release OperaLQ, a benchmark of real-world degraded opera videos. On this dataset, our method achieves state-of-the-art performance across multiple image and video quality metrics.
\end{itemize}
\section{Related Work}
\label{sec:related work}
\subsection{Video Super-Resolution}
Video super-resolution (VSR) aims to reconstruct HR frames from their low-resolution LR counterparts by exploiting spatial and temporal correlations across frames \cite{chan2021basicvsr}. Early approaches estimate optical flow for motion compensation to align neighboring frames before fusion \cite{kim2018spatio,xue2019video,sajjadi2018frame}, but motion estimation errors and occlusions often limit their accuracy. To overcome this, deformable convolution–based methods perform implicit alignment and aggregation in a data-driven manner \cite{tian2020tdan,jo2018deep}, and have been extended for intermediate-frame prediction \cite{wang2019edvr} or joint reconstruction within recurrent frameworks \cite{chan2022basicvsr++,chan2021understanding}. Recently, attention-driven architectures have been introduced to capture long-range temporal dependencies and enhance global feature interaction \cite{liang2024vrt,cao2021video,liang2022recurrent,shi2022rethinking}. Despite these advances, most VSR models are trained under simplified synthetic degradations (e.g., bicubic downsampling), leading to poor generalization to real-world scenarios with complex noise, compression, and motion blur.


\begin{figure*}[!t]
  \centering
   \includegraphics[width=1\linewidth]{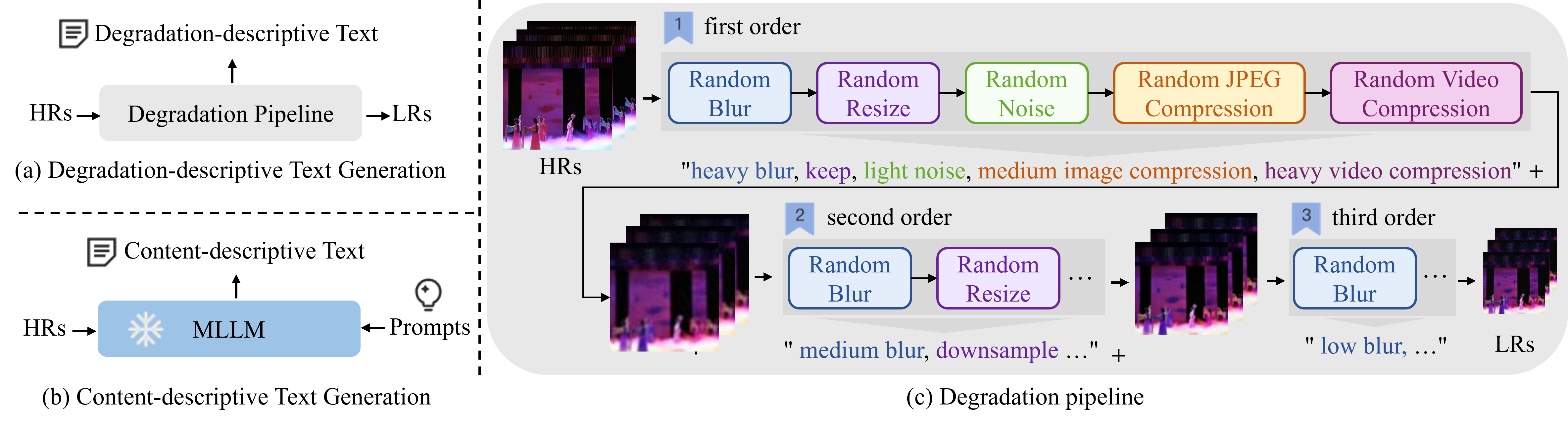}
   \caption{\textbf{Generation process of degradation- and content-descriptive texts.} Degradation-descriptive text is generated according to different intensity levels in the high-order degradation pipeline, while content-descriptive text is produced directly from high-resolution inputs (HRs) using a multimodal large language model (MLLM), rather than from degraded low-resolution videos (LRs).}
   \label{text_gen}
\end{figure*}

\subsection{Real-World Video Super-Resolution}
To address complex and unknown degradations in real-world videos, recent studies have explored constructing realistic LR–HR pairs through either physical acquisition or synthetic degradation modeling. Data collection–based approaches such as RealVSR \cite{yang2021real} employ synchronized multi-camera setups to capture paired sequences, but their dependence on specialized hardware limits scalability and general applicability. In contrast, degradation simulation–based methods focus on learning or expanding the degradation space. DBVSR \cite{pan2021deep} models blur kernels via convolutional learning, while AnimeSR \cite{wu2022animesr} broadens the degradation domain using diverse synthetic operators. Real-ESRGAN \cite{wang2021real} integrates multiple known kernels into a high-order degradation pipeline to better approximate real-world scenarios. NegVSR \cite{song2024negvsr} further enhances realism by sampling noise patterns directly from real data. 
Beyond degradation modeling, several works incorporate degradation correction into the network architecture. RealBasicVSR \cite{chan2022investigating} employs a dynamic cleaning module to suppress artifact propagation, while FastRealVSR \cite{xie2023mitigating} fuses sharpening and blur-kernel filtering for efficient compensation during hidden-state interaction. RealViformer \cite{zhang2024realviformer} further highlights that channel attention can mitigate redundant information and enhance robustness against residual artifacts. Despite these advancements, existing methods still face two key challenges: the degradation space remains incomplete or domain-specific, and the reliance on image-only features limits the recovery of high-quality videos with clear structures and natural textures.




\subsection{Text-guided Video Super-Resolution}

Recently, text-guided image super-resolution (ISR) has gained increasing attention \cite{wei2025perceive}. With powerful generative priors, text-to-image (T2I) diffusion models have shown strong potential for real-world SR \cite{wu2024seesr,yang2024pixel}, yet their multi-step denoising leads to high computational cost. To address this, several studies adopt knowledge distillation to achieve single-step diffusion \cite{wu2024one,chen2025adversarial,dong2025tsd}, effectively reducing complexity while maintaining perceptual quality. CLIP-SR \cite{hu2025clip} further integrates textual semantics into the reconstruction network to enhance fine texture recovery.
Building on these advances, recent works extend generative priors to video SR, exploring different strategies to alleviate fidelity degradation and temporal inconsistency caused by diffusion randomness \cite{zhou2024upscale,li2025diffvsr}. STAR \cite{xie2025star} further employs a text-to-video (T2V) diffusion model to enforce temporal coherence during generation. While diffusion-based approaches improve texture realism, their substantial computational overhead \cite{yang2024motion} still limits real-world deployment.
Different from these diffusion-based methods, our approach embeds multiple types of textual prompts into a classical RWVSR framework, enabling effective degradation modeling and texture enhancement in a lightweight and robust manner.

\section{Method}
\label{sec:method}

\begin{figure*}[!htbp]
  \centering
   \includegraphics[width=1\linewidth]{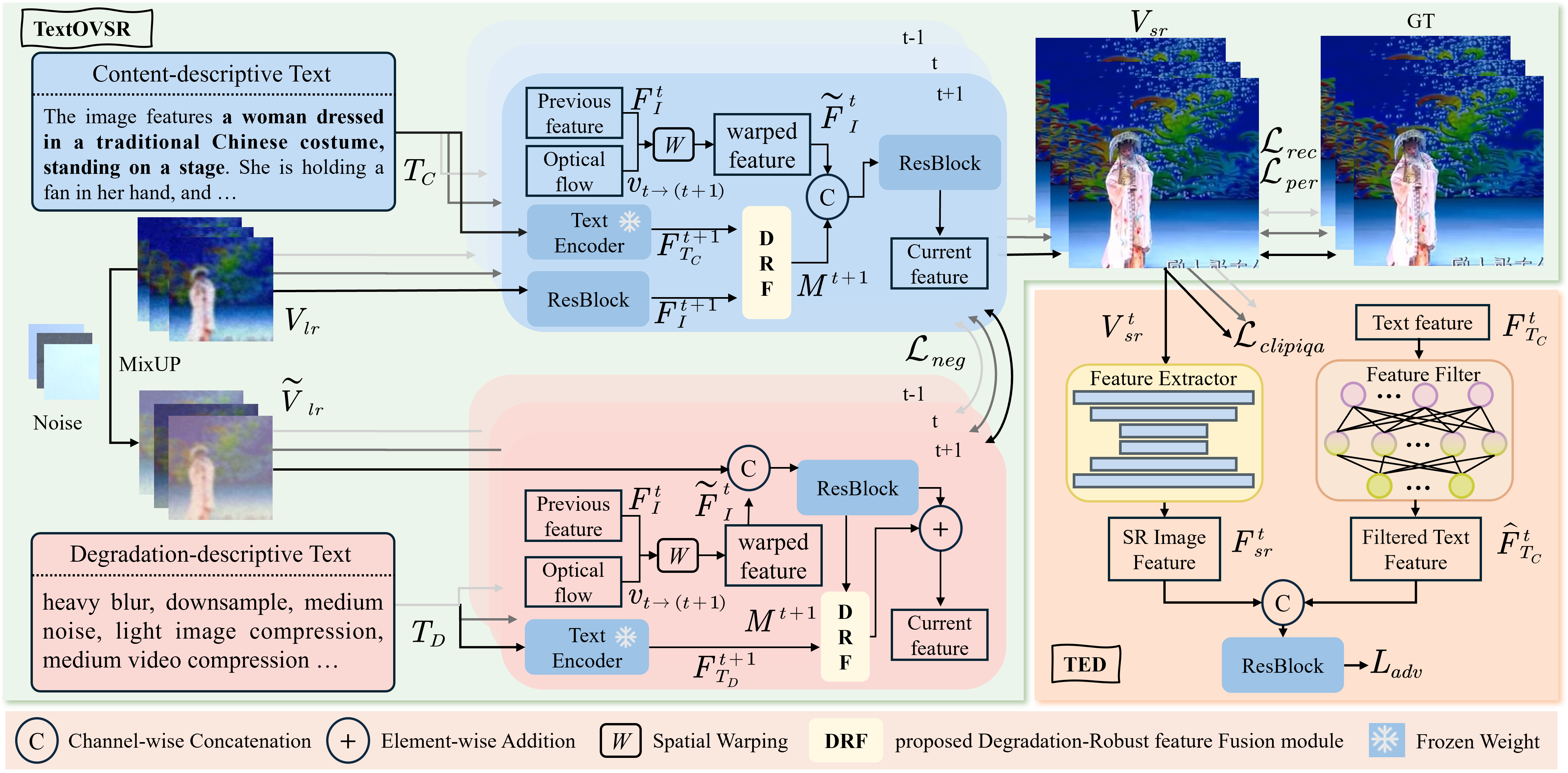}
   \caption{\textbf{The proposed TextOVSR network and TED adopt a two-stage training scheme.} In the first stage, only TextOVSR is trained. The positive branch (blue) takes content-descriptive text ($T_C$) and degraded videos ($V_{lr}$) as input, while the negative branch (red) takes degradation-descriptive text ($T_D$) and mixed-noise videos ($\widetilde{V}_{lr}$). Text features are extracted using a frozen CLIP encoder and fused with image features through the proposed DRF module. In the second stage, TextOVSR serves as the generator and TED as the discriminator. Adversarial training refines texture realism by selecting reliable textual features and integrating them with reconstructed image features. Here, $t-1$, $t$, and $t+1$ denote three consecutive frames, with detailed propagation described in \Cref{TextOVSR}.}
   \label{framework}
\end{figure*}
\subsection{Description Text Generation}
\label{Description Text Generation}

To provide textual guidance, we first produce degradation-descriptive text while synthesizing low-resolution (LR) videos using a high-order degradation pipeline \cite{wang2021real}, as illustrated in \Cref{text_gen}(a) and (c). Each stage of the pipeline includes operations such as blur, resize, noise, JPEG compression, and video compression. Following PromptSR \cite{chen2023image}, degradations are categorized into three intensity levels: light, medium, and heavy, yielding descriptions such as ``light blur''. For higher-order degradations, descriptions from successive stages are concatenated to form a comprehensive high-order degradation description.
For frame-level semantic guidance, a multi-modal large language model (MLLM) generates content-descriptive text for each frame, as shown in \Cref{text_gen}(b). This text captures the key visual semantics of each frame, providing high-level guidance for super-resolution. To improve efficiency and facilitate video-level degradation modeling, text is shared across consecutive frames in batches, with a batch size of seven to match the format of the original dataset.
During inference, only the positive branch is used. The inputs consist of the degraded video and its frame-wise content-descriptive text, while degradation-descriptive text is no longer required. The MLLM is used to generate content text for each test frame. The intensity levels of different degradation kernels and the MLLM prompts are provided in Appendix Section 1.

\subsection{TextOVSR}
\label{TextOVSR}
The proposed Text-guided Dual-Branch Opera Video Super-Resolution (TextOVSR) network comprises two branches, as illustrated in \Cref{framework}. The positive branch (blue) takes content-descriptive text and degraded low-resolution videos as input, producing the super-resolution output. The negative branch (red) takes degradation-descriptive text and low-resolution videos with noise as input. During training, outputs from both branches are used to compute the negative loss ($\mathcal{L}_{neg}$), enhancing the positive branch's robustness to real-world noise. During inference, only the positive branch is employed. Both branches are built on the BasicVSR architecture \cite{chan2021basicvsr}, with text features extracted via the CLIP text encoder \cite{radford2021learning} and fused with image features using the Degradation-Robust feature Fusion (DRF) module. The key distinction lies in the timing of image-text fusion. In the positive branch, text features are fused early, before deep feature extraction, enhancing frame feature expressiveness and mitigating error propagation. In the negative branch, text features are fused after deep feature extraction, allowing degradation descriptions to model real-world noise at the feature level. Ablation results on the fusion position are provided in \Cref{Impact of DRF Module Location}. TextOVSR is trained in two stages. In the first stage, TextOVSR is trained alone. In the second stage, the trained model serves as a generator, and the Text-Enhanced Discriminator (TED) is introduced to further improve reconstruction quality. Detailed training procedures are described in \Cref{Implementation Details}.

Both the positive and negative branches adopt a bidirectional propagation mechanism. Taking the forward propagation from time step $t$ to $t+1$ in the positive branch as an example, the process is as follows. The degraded video $V_{lr}\in \mathbb{R}^{n\times c\times h\times w}$ is first processed by a residual module to extract the feature of the frame at time step $t+1$, denoted as $F_{I}^{t+1}$. The pre-trained CLIP text encoder then encodes the content description $T_C$ of each frame into text feature $F_{T_C}^{t+1}$, which is fused with the corresponding frame feature via the DRF module to obtain the fused feature $M^{t+1}$. Meanwhile, the propagated feature $F_{I}^{t}$ and optical flow $v_{t\rightarrow t+1}$ are used to generate the aligned feature $\widetilde{F}_{I}^{t}$ through spatial warping. Finally, the fused feature and the aligned temporal feature are concatenated along the channel dimension, and a residual module produces the final feature for time step $t+1$.


\begin{figure}[!t]
  \centering
   \includegraphics[width=0.95\linewidth]{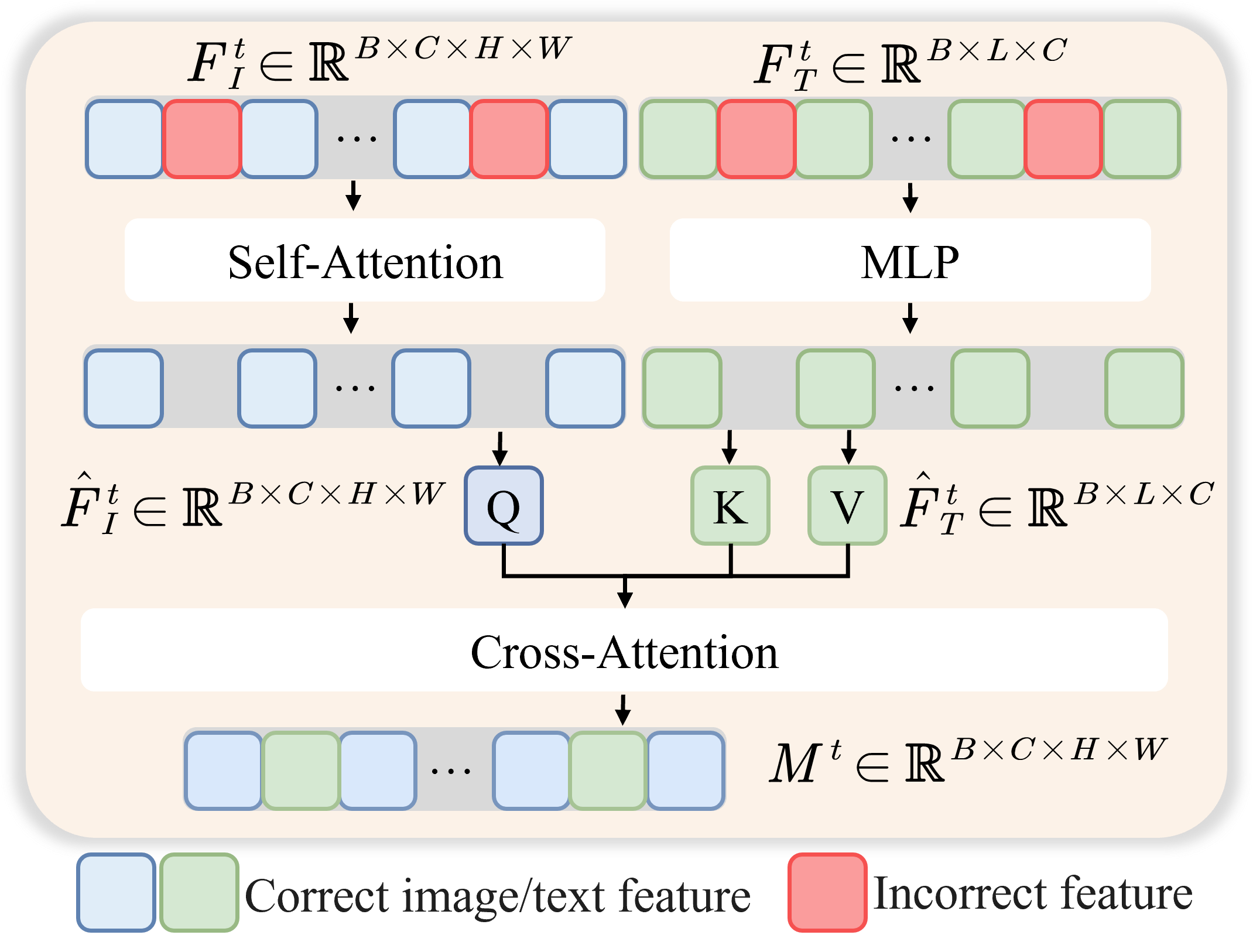}
   \caption{\textbf{The proposed Degradation-Robust Feature Fusion (DRF) module.} $F_{I}^{t+1}$ and $F_{T}^{t+1}$ denote the image and text feature vectors, respectively, and $M^{t}$ represents the fused feature. }
   \label{drf}
\end{figure}

\subsection{Degradation-Robust feature Fusion Module}
Input features from both positive and negative branches may contain unreliable information. In the positive branch, low-resolution inputs degraded through various processes inherently include distortions and erroneous features. Direct reliance on these features amplifies errors during temporal propagation, leading to blurred details and artifacts in reconstruction results \cite{chan2022investigating,xie2023mitigating}. To better simulate real-world degradation, noise extracted from external datasets is introduced in the negative branch. However, such noise often exhibits style mismatches. For example, noise from everyday videos may not match the stylistic characteristics of opera videos, resulting in abnormal artifacts, as as indicated by the yellow arrows in \Cref{fig1}. Furthermore, content descriptions generated by the MLLM may also contain inaccuracies, introducing an additional source of uncertainty.

To address inaccuracies in the input features, we design a Degradation-Robust Feature Fusion (DRF) module for image-text fusion, as illustrated in \Cref{drf}. The extracted frame features $F_{I}^{t+1}$ and text features $F_{T}^{t+1}$ are first processed by multi-head self-attention and linear layers, respectively, to amplify reliable information while suppressing noise and erroneous features. The filtered image feature $\hat{F}_{I}^{t}$ is then used to generate the Query (Q), and the filtered text feature $\hat{F}_{T}^{t}$ is used to generate the Key (K) and Value (V). Finally, a multi-head cross-attention mechanism computes the fused feature $M^{t+1}$.


\subsection{Text-Enhanced Discriminator}
The content-descriptive text contains high-level semantic information, which can enhance discrimination accuracy \cite{tao2023galip}. To leverage this, we introduce textual features into a standard UNet-based discriminator, forming the Text-Enhanced Discriminator (TED), as illustrated in \Cref{framework}. Specifically, the inputs include the super-resolved frame $V_{sr}^{t}$ and corresponding content description text feature $F_{T_C}^{t}$. A standard UNet extracts the image feature $F_{sr}^{t}$, while a feature filter emphasizes the effective textual features $\hat{F}_{T_C}^{t}$. The extracted image feature and filtered text feature are then concatenated along the channel dimension and processed by a residual module to compute the adversarial loss. This design fully exploits the high-level semantics from the content descriptions while mitigating the influence of inaccurate features, leading to more precise discrimination.

\subsection{Objective Functions}
To comprehensively train the proposed TextOVSR network, we adopt a two-stage training strategy following RealBasicVSR \cite{chan2022investigating} and NegVSR \cite{song2024negvsr}. In the first stage, TextOVSR is trained with the reconstruction loss ($\mathcal{L}_{rec}$) and the negative loss ($\mathcal{L}_{neg}$), formulated as:
\begin{equation}
\mathcal{L}_{stage1}=\mathcal{L}_{rec}\left( V_{sr}^{t},V_{GT}^{t} \right) +\alpha \mathcal{L}_{neg}\left( V_{sr}^{t},\hat{V}_{sr}^{t} \right)  ,
\end{equation}
where $V_{sr}^{t}$ and $\hat{V}_{sr}^{t}$ denote the outputs of the positive and negative branches, respectively. The loss weight $\alpha$ for $\mathcal{L}_{neg}$ is set to 0.5 following previous work \cite{song2024negvsr}.

In the second stage, TextOVSR serves as the generator, while the proposed TED acts as the discriminator. In addition to the reconstruction and negative losses from the first stage, the generator is further optimized with a perceptual loss ($\mathcal{L}_{per}$) \cite{johnson2016perceptual} and a CLIPIQA loss \cite{wang2023exploring} to enhance perceptual quality. The CLIPIQA loss is defined as:
\begin{equation}
\mathcal{L}_{clipiqa}=1-\mathcal{R}\left( V_{sr}^{t} \right) ,
\end{equation}
where $\mathcal{R}$ denotes the CLIP-IQA model. To further improve detail recovery, the adversarial loss is computed using TED:
\begin{equation}
\mathcal{L}_{adv}=-\mathbb{E}_{\left( \mathcal{H}\left( V_{lr}^{t},T_{C}^{t} \right) \sim P_g \right)}log\left( TED\left( V_{sr}^{t},F_{T_C}^{t} \right) \right) ,
\end{equation}
where $TED$ denotes the proposed text-enhanced discriminator, and $F_{T_C}^{t}$ represents the text features.The overall objective for stage two is:
\begin{equation}
\mathcal{L}_{stage2}=\mathcal{L}_{stage1}+\mathcal{L}_{per}\left( V_{sr}^{t},V_{GT}^{t} \right) +\beta \mathcal{L}_{clipiqa}+\mathcal{L}_{adv},
\end{equation}
where the hyperparameter $\beta$ is used to adjust the weight of the $\mathcal{L}_{clipiqa}$, which is set to 0.5 during training.
\section{Experiments}
\label{sec:experiments}

\begin{figure}[!t]
  \centering
   \includegraphics[width=0.9\linewidth]{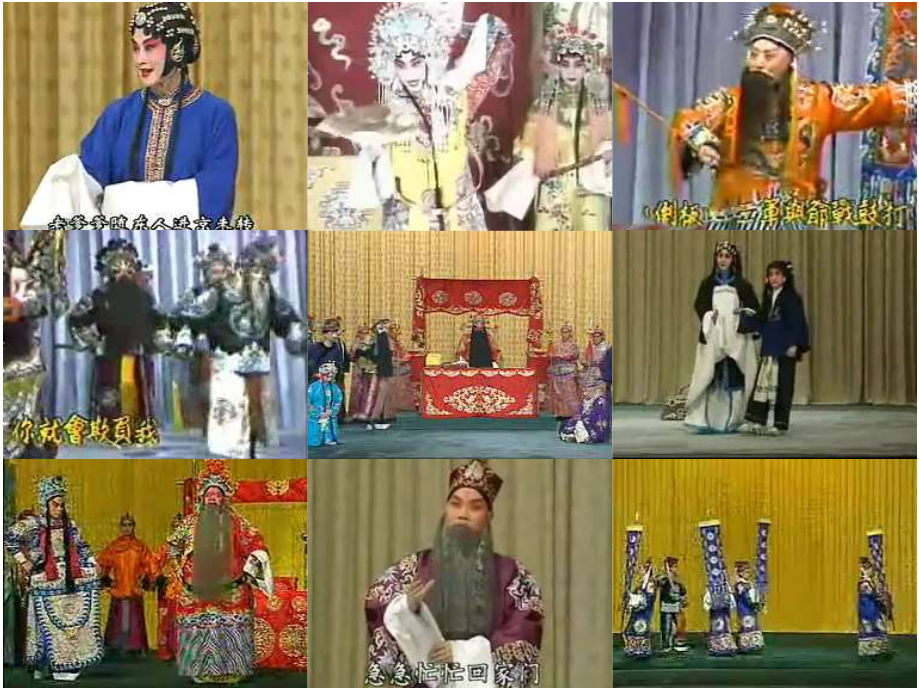}
   \caption{\textbf{OperaLQ Dataset.} Our OperaLQ dataset consists of real degraded opera videos with varying content and resolutions.}
   \label{operalq}\vspace*{-4mm}
\end{figure}
\subsection{Datasets and Metrics}

\begin{table*}[!t]
  \centering
  \begin{adjustbox}{max width=\textwidth}
  \begin{tblr}{
    width = \linewidth,
    colspec = {lccccccccc},  
    hline{1,14} = {1pt},
    hline{2,5} = {0.5pt}, 
    vline{2,3,4,5,6,7,8,9,10} = {0.5pt}, 
  }
    &Bicubic&DBVSR\cite{pan2021deep}&RealBasicVSR\cite{chan2022investigating}&FTVSR\cite{qiu2023learning}&Self-BlindVSR\cite{bai2024self}&RealViformer\cite{zhang2024realviformer}&NegVSR\cite{song2024negvsr}&BVSR-IK\cite{zhu2025blind}&Ours\\
    \makecell[l]{Params(M)}&- &36.3 &4.9 &45.8 &139.5 &8.5 &3.4 &6.5 &5.7 \\
    Runtimes(ms)&- &411.4 &81.2 &825.4 &985.6 &87.9 &63.8 &145.3 & 195.3\\
    FLOPs(G)&- &1159.2 &376.9 &2417.7 &2518.0 &213.0 &292.0 &317.4 & 309.6\\ 

    NRQM$\uparrow$&2.9193 &3.9296 &5.1708 &3.0731 &3.0474 &5.1894 &5.7761 &3.6682 & \textbf{5.8184} \\
    MUSIQ$\uparrow$&27.0694 &38.8246 &48.3548 &28.5172 &31.6829 &52.3948 &\textbf{58.6386} &39.6988 & 58.3033 \\
    CLIPIQA+$\uparrow$&0.4212 &0.3529 &0.3494 &0.2917 &0.3147 &0.3774 &0.3990 &0.3855 & \textbf{0.5667}\\
    TOPIQ$\uparrow$&0.2125 &0.2625 &0.3556 &0.2095 &0.2134 &0.3669 &0.4354 &0.2607 & \textbf{0.4636}\\
    BRISQUE$\downarrow$&61.3265 &56.2738 &41.7475 &53.7206 &61.8668 &39.3883 &33.5291 &57.9673 & \textbf{33.3799}\\
    NIQE$\downarrow$&7.5734 &6.0774 &4.4300 &7.0509 &6.9784 &4.4347 &4.0756 &6.8368 & \textbf{3.5139}\\
    ILNIQE$\downarrow$&37.9113 &29.2702 &31.1341 &35.4652 &34.0444 &32.7323 &32.4800 &33.7224 & \textbf{30.0242}\\
    PI$\downarrow$&7.3350 &6.2133 &4.7243 &7.0915 &7.0388 &4.7117 &4.2232 &6.6727 & \textbf{3.9913}\\
    DOVER$\uparrow$&8.6303 &15.9180 &33.4799 &9.5426 &11.8134 &39.4318 &40.6763 &16.7777 &\textbf{45.0415} \\
  \end{tblr}
  \end{adjustbox}
  \caption{\textbf{Quantitative comparison with existing methods.} Best and second-best results are highlighted. Model parameters (Params), runtime, and FLOPs are evaluated on the same device. $\uparrow$ denotes higher values are better; $\downarrow$ denotes lower values are better.}
  \label{Quantitative comparisons}
\end{table*}

\begin{figure*}[!htbp]
  \centering
   \includegraphics[width=1\linewidth]{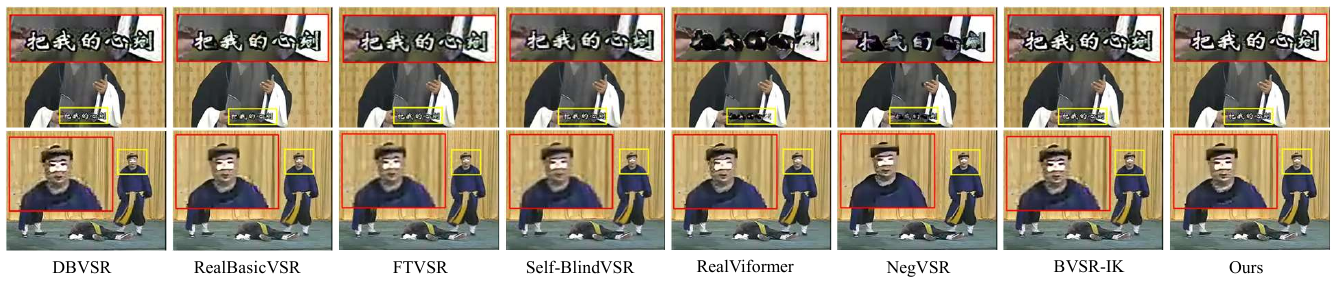}
    \caption{\textbf{Qualitative comparison with other methods.} (Zoom-in for best view.)}
    \label{Qualitative comparison}
\end{figure*}



\textbf{Dataset.} 
For training, we use the Chinese Opera Video Clips (COVC) dataset introduced in MambaOVSR \cite{chang2025mambaovsr}. Each 7-frame clip is split into individual frames for reconstruction. Unlike RealBasicVSR \cite{chan2022investigating}, which applies online degradations, we pre-generate degraded inputs using the RealESRGAN \cite{wang2021real} pipeline to ensure consistent degradations across epochs and accurate alignment with the degradation-descriptive text for each frame. As described in \Cref{Description Text Generation}, content-descriptive texts are produced by an MLLM, while degradation-descriptive texts are obtained through a binning strategy. Detailed parameter settings and classification criteria are provided in the appendix. During preprocessing, the ground-truth (GT) frames and their degraded counterparts are randomly cropped to $256 \times 256$. The degraded frames are then downsampled to $64 \times 64$ using bicubic interpolation. Random horizontal flipping is applied for augmentation. For evaluation, we construct a real-world benchmark named OperaLQ, consisting of 50 opera videos with 100 frames each. As shown in \Cref{operalq}, the videos are collected from diverse sources to cover a broad range of real-world degradations, including various motion patterns, resolutions, and scene complexities.

\textbf{Metric.}
Since the test set contains real-world degraded videos without ground-truth references, reference-based metrics are strictly inapplicable. To comprehensively and more robustly evaluate overall real-world video super-resolution performance, we employ a suite of reference-free image and video quality metrics. For image-level evaluation, we adopt NRQM \cite{ma2017learning}, MUSIQ \cite{ke2021musiq}, CLIPIQA+ \cite{wang2023exploring}, TOPIQ \cite{chen2024topiq}, BRISQUE \cite{mittal2011blind}, NIQE \cite{mittal2012making}, ILNIQE \cite{zhang2015feature}, and PI \cite{blau20182018} to measure perceptual quality and naturalness. For video-level evaluation, we use DOVER \cite{wu2023exploring} to assess overall temporal and perceptual consistency.

\subsection{Implementation Details}
\label{Implementation Details}
We use the pre-trained SPyNet \cite{ranjan2017optical} to estimate optical flow between adjacent frames, with its weights frozen during training. The content-descriptive texts are generated by LLaVA \cite{liu2023visual}, and their textual embeddings are extracted using the CLIP text encoder \cite{radford2021learning} with a ViT-L/14@336px backbone. The training process is divided into two stages. In the first stage, the TextOVSR model is trained for 100K iterations using the Adam optimizer with a learning rate of $1\times10^{-4}$ and the loss function $\mathcal{L}_{stage1}$. In the second stage, the model is fine-tuned within a GAN framework to enhance fine details and perceptual realism, using a reduced learning rate of $5\times10^{-5}$ and the loss function $\mathcal{L}_{stage2}$.

\begin{table*}[!t]
  \centering
  \resizebox{\linewidth}{!}{
  \begin{tblr}{
    width = \linewidth,
    colspec = {cccccccccccc},  
    hline{1,9} = {1pt},
    hline{2,3,4,5,6,7,8} = {0.5pt}, 
    vline{2,3,4,5,6,7,8} = {0.5pt}, 
  }
  \SetCell[r=2]{c} Variants& \SetCell[r=2]{c} baseline& \SetCell[c=4]{c} DRF&&&& \SetCell[r=2]{c} TED&\SetCell[r=2]{c}NRQM$\uparrow$ &\SetCell[r=2]{c}CLIPIQA+$\uparrow$&\SetCell[r=2]{c}TOPIQ$\uparrow$&\SetCell[r=2]{c}NIQE$\downarrow$&\SetCell[r=2]{c}BRISQUE$\downarrow$\\
  & & w/o $T,N$&w/o $T$& w $T_D$ & w $T_D\&T_C$& & & \\
  1 & \usym{2713}&\usym{2717}&\usym{2717}&\usym{2717}&\usym{2717}&\usym{2717}& 5.7761 & 0.3990& 0.4354& 4.0756& 33.5291\\
  2 & \usym{2713}&\usym{2713}&\usym{2717}&\usym{2717}&\usym{2717}&\usym{2717}& 5.4949 & 0.5462& 0.4436& 3.7303& 38.3618\\
  3 & \usym{2713}&\usym{2717}&\usym{2717}&\usym{2713}&\usym{2717}&\usym{2717}& 5.4610 & 0.5471& 0.4483& 3.6396& 39.3594\\
  4 & \usym{2713}&\usym{2717}&\usym{2713}&\usym{2717}&\usym{2717}&\usym{2717}& 5.5697 & 0.5507& 0.4523& 3.6627& 38.1570\\
  5 & \usym{2713}&\usym{2717}&\usym{2717}&\usym{2717}&\usym{2713}&\usym{2717}& 5.6838 & \textbf{0.5667}& 0.4636& 3.5139& 35.7444\\
  6(Ours) & \usym{2713}&\usym{2717}&\usym{2717}&\usym{2717}&\usym{2713}&\usym{2713}& \textbf{5.8184} & 0.5659& \textbf{0.4733}&\textbf{3.4291}& \textbf{33.3799}\\
  \end{tblr}
  }
  \caption{\textbf{Comparison of quantitative results with different components of our method on the OperaLQ dataset.} w/o $T,N$ denotes enhancing only the negative branch propagation, while w/o $T$ denotes enhancing both positive and negative branches simultaneously. w $T_D$ and w $T_D\&T_C$ denote using only degradation-descriptive text and using both degradation-descriptive and content-descriptive texts.}
  \label{ablation}
\end{table*}

\begin{figure*}[!t]
  \centering
   \includegraphics[width=1\linewidth]{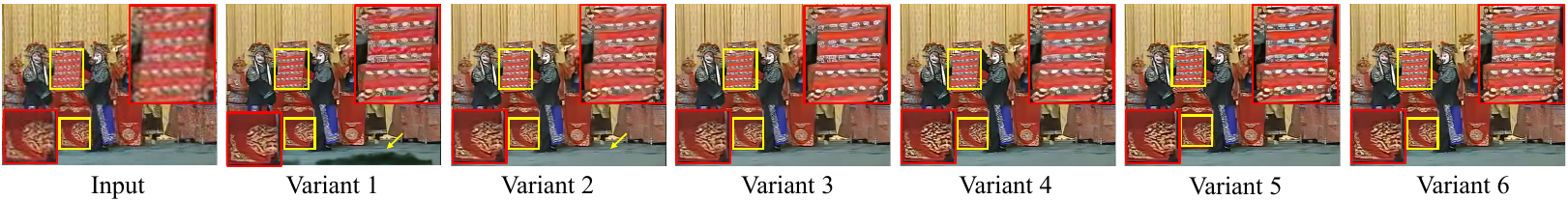}
   \caption{\textbf{Qualitative comparison of different component ablation studies.} (Zoom-in for best view.)}
   \label{ablation_Quali}
\end{figure*}

\subsection{Comparison with State-of-the-Arts}
We compare the proposed method with all publicly reproducible real-world video super-resolution (VSR) approaches, including Self-BlindVSR \cite{bai2024self}, NegVSR \cite{song2024negvsr}, RealViformer \cite{zhang2024realviformer}, FTVSR \cite{qiu2023learning}, RealBasicVSR \cite{chan2022investigating}, and DBVSR \cite{pan2021deep}. In addition, we evaluate against BVSR-IK \cite{zhu2025blind}, a state-of-the-art framework that jointly performs video super-resolution and deblurring.


The quantitative results on the OperaLQ dataset are shown in \Cref{Quantitative comparisons}. Compared with existing methods, TextOVSR achieves superior performance. Specifically, it attains the best results on multiple image quality metrics (e.g., CLIPIQA+, TOPIQ, NIQE) and achieves the highest score on the video metric DOVER. Qualitative results in \Cref{Qualitative comparison} show that TextOVSR better suppresses blur and restores fine details, especially in text and facial regions, yielding clearer and more faithful visual results than other methods. 


%

\section{Analysis and Discussions}
\label{sec:discussion}
\subsection{Ablation Study}
\label{Ablation Study}
We conducted ablation studies on OperaLQ to evaluate each component. Using NegVSR \cite{song2024negvsr} as the baseline (Variant 1), we added the DRF module to enhance dual-branch feature propagation with textual information. Four variants were then created to analyze individual contributions: enhancing only the negative branch, fusing only degradation text, enhancing both branches, and fusing both texts (Variants 2–5). Adding TED to Variant 5 yielded Variant 6.

Quantitative and qualitative results are shown in \Cref{ablation} and \Cref{ablation_Quali}. Variant 2 enhances the negative branch via the DRF module, effectively suppressing artifacts caused by style-inconsistent noise in Variant 1 (as indicated by arrows) and improving CLIPIQA+ by 0.1472. Variant 3 further incorporates degradation-descriptive text, yielding additional improvements in CLIPIQA+. Variant 4 strengthens both positive and negative branches, reducing erroneous features in the positive branch and producing sharper contours. Variant 5 integrates degradation- and content-descriptive texts, significantly enhancing texture representation. Finally, the introduction of TED to guide texture reconstruction further boosts overall reconstruction quality.

\subsection{Impacts of Coarse-Grained and Fine-Grained Content-descriptive text}
\label{Impact of Coarse-Grained and Fine-Grained Content-descriptive text}
\begin{figure}[!t]
  \centering
   \includegraphics[width=1\linewidth]{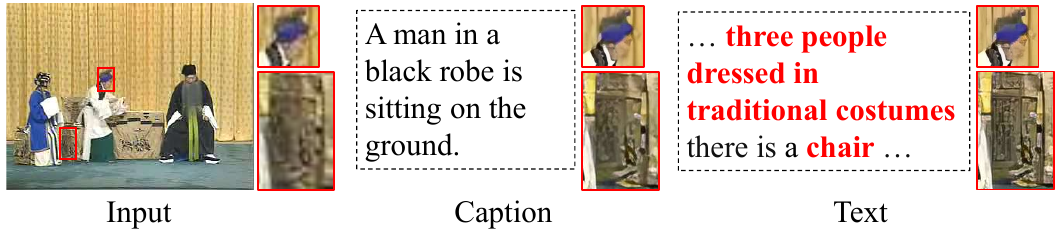}
   \caption{\textbf{Qualitative comparison of coarse-grained and fine-grained textual descriptions.} (Zoom-in for best view.)}
   \label{ablation_text}
\end{figure}

To study the effect of textual granularity on super-resolution, we generated two types of content descriptions for the OperaLQ test set: Caption (coarse-grained) and Text (fine-grained). Quantitative results (\Cref{analysis of coarse-grained and fine-grained textual descriptions}) show that fine-grained descriptions improve overall reconstruction with only a minor CLIPIQA+ drop (0.0051). Qualitative results (\Cref{ablation_text}) indicate that fine-grained texts better guide recovery of fine structures, producing clearer and more realistic textures (e.g., facial and chair regions).

\begin{table}[!t]
  \centering
  \resizebox{\linewidth}{!}{
  \begin{tblr}{
    width = \linewidth,
    colspec = {ccccc},  
    hline{1,4} = {1pt},
    hline{2,3} = {0.5pt}, 
    vline{2} = {0.5pt},
  }
  &NRQM$\uparrow$& CLIPIQA+$\uparrow$& TOPIQ$\uparrow$&NIQE$\downarrow$\\
  Caption&5.5552& \textbf{0.5718}& 0.4602& 3.6206\\
  Text&\textbf{5.6838}& 0.5667& \textbf{0.4636}& \textbf{3.5139}\\
  \end{tblr}
  }
  \caption{\textbf{Quantitative comparison of coarse-grained and fine-grained textual descriptions}, where Caption denotes coarse-grained textual descriptions and Text denotes fine-grained ones.}
  \label{analysis of coarse-grained and fine-grained textual descriptions}
\end{table}

\subsection{Impacts of Different Discriminators}
\label{Impact of Different Discriminators}
\begin{figure}[!t]
  \centering
   \includegraphics[width=1\linewidth]{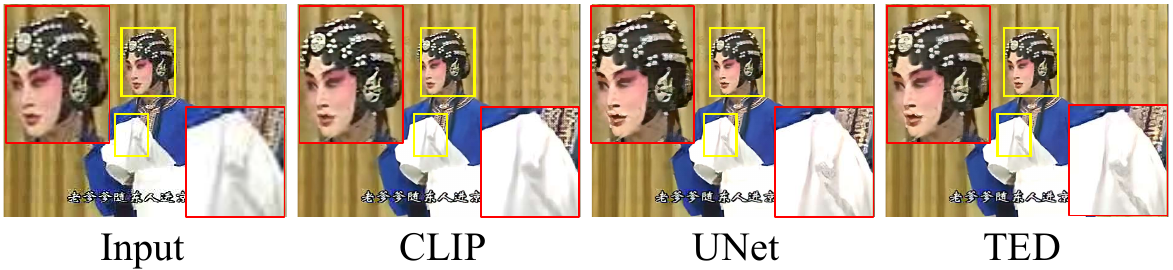}
   \caption{\textbf{Qualitative comparison of different Discriminators.} (Zoom-in for best view.)}
   \label{ablation_d}
\end{figure}

\Cref{analysis of different discriminators} and \Cref{ablation_d} present quantitative and qualitative comparisons of different discriminator architectures. The UNet-based discriminator (UNet) remains the most commonly adopted architecture in previous works \cite{chan2022investigating,song2024negvsr}. GALIP \cite{tao2023galip} directly utilizes CLIP’s image and text encoders for feature extraction and alignment (CLIP); however, its unfiltered alignment often leads to ambiguous reconstructions, particularly in fine-grained regions such as faces and sleeves. In contrast, our proposed TED employs a UNet for image feature extraction while selectively filtering textual features, producing more accurate and sharper reconstructions and delivering the best overall performance.



\begin{table}[!t]
  \centering
  \resizebox{\linewidth}{!}{
  \begin{tblr}{
    width = \linewidth,
    colspec = {ccccc},  
    hline{1,5} = {1pt},
    hline{2,3,4} = {0.5pt}, 
    vline{2} = {0.5pt},
  }
  &NRQM$\uparrow$& CLIPIQA+$\uparrow$& TOPIQ$\uparrow$&NIQE$\downarrow$\\
  CLIP&5.0118& 0.3081& 0.2860& 5.3520 \\
  UNet&5.6838& \textbf{0.5667}& 0.4636& 3.5139\\
  TED&\textbf{5.8184}& 0.5659& \textbf{0.4733}& \textbf{3.4291}\\
  \end{tblr}
  }
  \caption{\textbf{Quantitative comparison of different discriminators.}}
  \label{analysis of different discriminators}
\end{table}

\subsection{Impacts of DRF Module Location}
\label{Impact of DRF Module Location}
\begin{figure}[!t]
  \centering
   \includegraphics[width=1\linewidth]{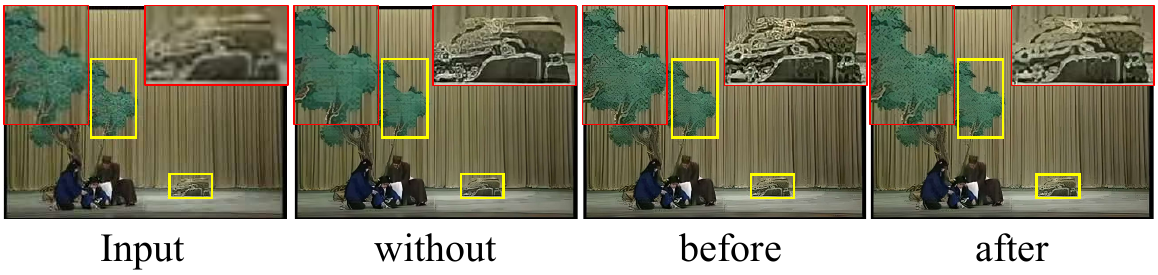}
   \caption{\textbf{Qualitative comparison of DRF at different positions in the negative branch.} (Zoom-in for best view.)}
   \label{ablation_drf_location}
\end{figure}

In the positive branch, DRF is placed before image feature extraction to fuse image and text features effectively. In the negative branch, we test DRF placement before or after feature extraction. Qualitative results (\Cref{ablation_drf_location}) show that pre-extraction fusion recovers richer details but adds noise, while post-extraction fusion suppresses out-of-distribution noise and improves quantitative metrics (\Cref{analysis of location of drf}).

\begin{table}[!t]
  \centering
   \resizebox{\linewidth}{!}{
    \begin{tblr}{
        colspec = {ccccc},  
        hline{1,5} = {1pt},
        hline{2,3,4} = {0.5pt}, 
        vline{2} = {0.5pt},
      }
      Location& NRQM$\uparrow$&CLIPIQA+$\uparrow$& TOPIQ$\uparrow$&NIQE$\downarrow$\\
      without& \textbf{5.7761}& 0.3990& 0.4354& 4.0753\\
      before& 5.4280& 0.5438& 0.4316 & \textbf{3.7274}\\
      after& 5.4949& \textbf{0.5462}& \textbf{0.4436}& 3.7303\
      \end{tblr}
   }
  \caption{\textbf{Quantitative comparison of DRF at different positions in the negative branch} (before and after image feature extraction). without indicates no DRF is added.}
  \label{analysis of location of drf}
\end{table}

\section{Conclusion}
\label{sec:conclusion}

In this work, we propose a novel Text-guided Dual-Branch Opera Video Super-Resolution (TextOVSR) network for real-world opera video super-resolution. Specifically, the negative branch of TextOVSR integrates degradation-descriptive text derived from the degradation process to broaden the degradation space. Simultaneously, to enhance texture reconstruction, content-descriptive text is incorporated into both the positive branch and the proposed TED. Furthermore, we design a DRF module that alleviates degradation-induced contamination while enabling effective cross-modal fusion of visual and textual features. On our constructed real-world benchmark, OperaLQ, TextOVSR outperforms existing state-of-the-art methods in both quantitative and qualitative evaluations.

\section{Acknowledgments}
This work was supported by the Natural Science Foundation of China (62376201, and 62501189), Hubei Provincial Science \& Technology Talent Enterprise Services Program (2025DJB059), Hubei Provincial Special Fund for Central-Guided Local S\&T Development (2025CSA017), and the Natural Science Foundation of Heilongjiang Province of China for Excellent Youth Project (YQ2024F006).
{
    \small
    \bibliographystyle{ieeenat_fullname}
    \bibliography{main}

\begin{thebibliography}{51}
\providecommand{\natexlab}[1]{#1}
\providecommand{\url}[1]{\texttt{#1}}
\expandafter\ifx\csname urlstyle\endcsname\relax
  \providecommand{\doi}[1]{doi: #1}\else
  \providecommand{\doi}{doi: \begingroup \urlstyle{rm}\Url}\fi

\bibitem[Bai and Pan(2024)]{bai2024self}
Haoran Bai and Jinshan Pan.
\newblock Self-supervised deep blind video super-resolution.
\newblock \emph{IEEE TPAMI}, 46\penalty0 (7):\penalty0 4641--4653, 2024.

\bibitem[Blau et~al.(2018)Blau, Mechrez, Timofte, Michaeli, and Zelnik-Manor]{blau20182018}
Yochai Blau, Roey Mechrez, Radu Timofte, Tomer Michaeli, and Lihi Zelnik-Manor.
\newblock The 2018 pirm challenge on perceptual image super-resolution.
\newblock In \emph{ECCVW}, pages 0--0, 2018.

\bibitem[Cao et~al.(2021)Cao, Li, Zhang, and Van~Gool]{cao2021video}
Jiezhang Cao, Yawei Li, Kai Zhang, and Luc Van~Gool.
\newblock Video super-resolution transformer.
\newblock \emph{arXiv preprint arXiv:2106.06847}, 2021.

\bibitem[Chan et~al.(2021{\natexlab{a}})Chan, Wang, Yu, Dong, and Loy]{chan2021basicvsr}
Kelvin~CK Chan, Xintao Wang, Ke Yu, Chao Dong, and Chen~Change Loy.
\newblock Basicvsr: The search for essential components in video super-resolution and beyond.
\newblock In \emph{CVPR}, pages 4947--4956, 2021{\natexlab{a}}.

\bibitem[Chan et~al.(2021{\natexlab{b}})Chan, Wang, Yu, Dong, and Loy]{chan2021understanding}
Kelvin~CK Chan, Xintao Wang, Ke Yu, Chao Dong, and Chen~Change Loy.
\newblock Understanding deformable alignment in video super-resolution.
\newblock In \emph{AAAI}, pages 973--981, 2021{\natexlab{b}}.

\bibitem[Chan et~al.(2022{\natexlab{a}})Chan, Zhou, Xu, and Loy]{chan2022basicvsr++}
Kelvin~CK Chan, Shangchen Zhou, Xiangyu Xu, and Chen~Change Loy.
\newblock Basicvsr++: Improving video super-resolution with enhanced propagation and alignment.
\newblock In \emph{CVPR}, pages 5972--5981, 2022{\natexlab{a}}.

\bibitem[Chan et~al.(2022{\natexlab{b}})Chan, Zhou, Xu, and Loy]{chan2022investigating}
Kelvin~CK Chan, Shangchen Zhou, Xiangyu Xu, and Chen~Change Loy.
\newblock Investigating tradeoffs in real-world video super-resolution.
\newblock In \emph{CVPR}, pages 5962--5971, 2022{\natexlab{b}}.

\bibitem[Chang et~al.(2025)Chang, Xu, Liu, Wang, Yuan, and Jiang]{chang2025mambaovsr}
Hua Chang, Xin Xu, Wei Liu, Wei Wang, Xin Yuan, and Kui Jiang.
\newblock Mambaovsr: Multiscale fusion with global motion modeling for chinese opera video super-resolution.
\newblock \emph{arXiv preprint arXiv:2511.06172}, 2025.

\bibitem[Chen et~al.(2025)Chen, Li, Wu, Zhang, Chen, Zhang, and Zhang]{chen2025adversarial}
Bin Chen, Gehui Li, Rongyuan Wu, Xindong Zhang, Jie Chen, Jian Zhang, and Lei Zhang.
\newblock Adversarial diffusion compression for real-world image super-resolution.
\newblock In \emph{CVPR}, pages 28208--28220, 2025.

\bibitem[Chen et~al.(2024)Chen, Mo, Hou, Wu, Liao, Sun, Yan, and Lin]{chen2024topiq}
Chaofeng Chen, Jiadi Mo, Jingwen Hou, Haoning Wu, Liang Liao, Wenxiu Sun, Qiong Yan, and Weisi Lin.
\newblock Topiq: A top-down approach from semantics to distortions for image quality assessment.
\newblock \emph{IEEE TIP}, 33:\penalty0 2404--2418, 2024.

\bibitem[Chen et~al.(2023)Chen, Zhang, Gu, Yuan, Kong, Chen, and Yang]{chen2023image}
Zheng Chen, Yulun Zhang, Jinjin Gu, Xin Yuan, Linghe Kong, Guihai Chen, and Xiaokang Yang.
\newblock Image super-resolution with text prompt diffusion.
\newblock \emph{arXiv preprint arXiv:2311.14282}, 2023.

\bibitem[Dong et~al.(2025)Dong, Fan, Guo, Wang, Zhang, Chen, Luo, and Zou]{dong2025tsd}
Linwei Dong, Qingnan Fan, Yihong Guo, Zhonghao Wang, Qi Zhang, Jinwei Chen, Yawei Luo, and Changqing Zou.
\newblock Tsd-sr: One-step diffusion with target score distillation for real-world image super-resolution.
\newblock In \emph{CVPR}, pages 23174--23184, 2025.

\bibitem[Hu et~al.(2025)Hu, Liu, Zheng, and Liu]{hu2025clip}
Bingwen Hu, Heng Liu, Zhedong Zheng, and Ping Liu.
\newblock Clip-sr: Collaborative linguistic and image processing for super-resolution.
\newblock \emph{IEEE TMM}, 2025.

\bibitem[Jo et~al.(2018)Jo, Oh, Kang, and Kim]{jo2018deep}
Younghyun Jo, Seoung~Wug Oh, Jaeyeon Kang, and Seon~Joo Kim.
\newblock Deep video super-resolution network using dynamic upsampling filters without explicit motion compensation.
\newblock In \emph{CVPR}, pages 3224--3232, 2018.

\bibitem[Johnson et~al.(2016)Johnson, Alahi, and Fei-Fei]{johnson2016perceptual}
Justin Johnson, Alexandre Alahi, and Li Fei-Fei.
\newblock Perceptual losses for real-time style transfer and super-resolution.
\newblock In \emph{ECCV}, pages 694--711. Springer, 2016.

\bibitem[Ke et~al.(2021)Ke, Wang, Wang, Milanfar, and Yang]{ke2021musiq}
Junjie Ke, Qifei Wang, Yilin Wang, Peyman Milanfar, and Feng Yang.
\newblock Musiq: Multi-scale image quality transformer.
\newblock In \emph{ICCV}, pages 5148--5157, 2021.

\bibitem[Kim et~al.(2018)Kim, Sajjadi, Hirsch, and Scholkopf]{kim2018spatio}
Tae~Hyun Kim, Mehdi~SM Sajjadi, Michael Hirsch, and Bernhard Scholkopf.
\newblock Spatio-temporal transformer network for video restoration.
\newblock In \emph{ECCV}, pages 106--122, 2018.

\bibitem[Li et~al.(2025)Li, Liu, Cao, Chen, Zhuang, Chen, He, Wang, and Qiao]{li2025diffvsr}
Xiaohui Li, Yihao Liu, Shuo Cao, Ziyan Chen, Shaobin Zhuang, Xiangyu Chen, Yinan He, Yi Wang, and Yu Qiao.
\newblock Diffvsr: Enhancing real-world video super-resolution with diffusion models for advanced visual quality and temporal consistency.
\newblock \emph{arXiv e-prints}, pages arXiv--2501, 2025.

\bibitem[Liang et~al.(2022)Liang, Fan, Xiang, Ranjan, Ilg, Green, Cao, Zhang, Timofte, and Gool]{liang2022recurrent}
Jingyun Liang, Yuchen Fan, Xiaoyu Xiang, Rakesh Ranjan, Eddy Ilg, Simon Green, Jiezhang Cao, Kai Zhang, Radu Timofte, and Luc~V Gool.
\newblock Recurrent video restoration transformer with guided deformable attention.
\newblock \emph{NeurIPS}, 35:\penalty0 378--393, 2022.

\bibitem[Liang et~al.(2024)Liang, Cao, Fan, Zhang, Ranjan, Li, Timofte, and Van~Gool]{liang2024vrt}
Jingyun Liang, Jiezhang Cao, Yuchen Fan, Kai Zhang, Rakesh Ranjan, Yawei Li, Radu Timofte, and Luc Van~Gool.
\newblock Vrt: A video restoration transformer.
\newblock \emph{IEEE TIP}, 33:\penalty0 2171--2182, 2024.

\bibitem[Liu et~al.(2023)Liu, Li, Wu, and Lee]{liu2023visual}
Haotian Liu, Chunyuan Li, Qingyang Wu, and Yong~Jae Lee.
\newblock Visual instruction tuning.
\newblock \emph{NeurIPS}, 36:\penalty0 34892--34916, 2023.

\bibitem[Ma et~al.(2017)Ma, Yang, Yang, and Yang]{ma2017learning}
Chao Ma, Chih-Yuan Yang, Xiaokang Yang, and Ming-Hsuan Yang.
\newblock Learning a no-reference quality metric for single-image super-resolution.
\newblock \emph{Computer Vision and Image Understanding}, 158:\penalty0 1--16, 2017.

\bibitem[Mittal et~al.(2011)Mittal, Moorthy, and Bovik]{mittal2011blind}
Anish Mittal, Anush~K Moorthy, and Alan~C Bovik.
\newblock Blind/referenceless image spatial quality evaluator.
\newblock In \emph{2011 conference record of the forty fifth asilomar conference on signals, systems and computers (ASILOMAR)}, pages 723--727. IEEE, 2011.

\bibitem[Mittal et~al.(2012)Mittal, Soundararajan, and Bovik]{mittal2012making}
Anish Mittal, Rajiv Soundararajan, and Alan~C Bovik.
\newblock Making a “completely blind” image quality analyzer.
\newblock \emph{IEEE Sign. Process. Letters}, 20\penalty0 (3):\penalty0 209--212, 2012.

\bibitem[Pan et~al.(2021)Pan, Bai, Dong, Zhang, and Tang]{pan2021deep}
Jinshan Pan, Haoran Bai, Jiangxin Dong, Jiawei Zhang, and Jinhui Tang.
\newblock Deep blind video super-resolution.
\newblock In \emph{ICCV}, pages 4811--4820, 2021.

\bibitem[Qiu et~al.(2023)Qiu, Yang, Fu, Liu, Xu, and Fu]{qiu2023learning}
Zhongwei Qiu, Huan Yang, Jianlong Fu, Daochang Liu, Chang Xu, and Dongmei Fu.
\newblock Learning degradation-robust spatiotemporal frequency-transformer for video super-resolution.
\newblock \emph{IEEE TPAMI}, 45\penalty0 (12):\penalty0 14888--14904, 2023.

\bibitem[Radford et~al.(2021)Radford, Kim, Hallacy, Ramesh, Goh, Agarwal, Sastry, Askell, Mishkin, Clark, et~al.]{radford2021learning}
Alec Radford, Jong~Wook Kim, Chris Hallacy, Aditya Ramesh, Gabriel Goh, Sandhini Agarwal, Girish Sastry, Amanda Askell, Pamela Mishkin, Jack Clark, et~al.
\newblock Learning transferable visual models from natural language supervision.
\newblock In \emph{ICML}, pages 8748--8763. PMLR, 2021.

\bibitem[Ranjan and Black(2017)]{ranjan2017optical}
Anurag Ranjan and Michael~J Black.
\newblock Optical flow estimation using a spatial pyramid network.
\newblock In \emph{CVPR}, pages 4161--4170, 2017.

\bibitem[Sajjadi et~al.(2018)Sajjadi, Vemulapalli, and Brown]{sajjadi2018frame}
Mehdi~SM Sajjadi, Raviteja Vemulapalli, and Matthew Brown.
\newblock Frame-recurrent video super-resolution.
\newblock In \emph{CVPR}, pages 6626--6634, 2018.

\bibitem[Shi et~al.(2022)Shi, Gu, Xie, Wang, Yang, and Dong]{shi2022rethinking}
Shuwei Shi, Jinjin Gu, Liangbin Xie, Xintao Wang, Yujiu Yang, and Chao Dong.
\newblock Rethinking alignment in video super-resolution transformers.
\newblock \emph{NeurIPS}, 35:\penalty0 36081--36093, 2022.

\bibitem[Song et~al.(2024)Song, Wang, Yang, Xian, and Shi]{song2024negvsr}
Yexing Song, Meilin Wang, Zhijing Yang, Xiaoyu Xian, and Yukai Shi.
\newblock Negvsr: Augmenting negatives for generalized noise modeling in real-world video super-resolution.
\newblock In \emph{AAAI}, pages 10705--10713, 2024.

\bibitem[Tao et~al.(2023)Tao, Bao, Tang, and Xu]{tao2023galip}
Ming Tao, Bing-Kun Bao, Hao Tang, and Changsheng Xu.
\newblock Galip: Generative adversarial clips for text-to-image synthesis.
\newblock In \emph{CVPR}, pages 14214--14223, 2023.

\bibitem[Tian et~al.(2020)Tian, Zhang, Fu, and Xu]{tian2020tdan}
Yapeng Tian, Yulun Zhang, Yun Fu, and Chenliang Xu.
\newblock Tdan: Temporally-deformable alignment network for video super-resolution.
\newblock In \emph{CVPR}, pages 3360--3369, 2020.

\bibitem[Wang et~al.(2023)Wang, Chan, and Loy]{wang2023exploring}
Jianyi Wang, Kelvin~CK Chan, and Chen~Change Loy.
\newblock Exploring clip for assessing the look and feel of images.
\newblock In \emph{AAAI}, pages 2555--2563, 2023.

\bibitem[Wang et~al.(2019)Wang, Chan, Yu, Dong, and Change~Loy]{wang2019edvr}
Xintao Wang, Kelvin~CK Chan, Ke Yu, Chao Dong, and Chen Change~Loy.
\newblock Edvr: Video restoration with enhanced deformable convolutional networks.
\newblock In \emph{CVPRW}, pages 0--0, 2019.

\bibitem[Wang et~al.(2021)Wang, Xie, Dong, and Shan]{wang2021real}
Xintao Wang, Liangbin Xie, Chao Dong, and Ying Shan.
\newblock Real-esrgan: Training real-world blind super-resolution with pure synthetic data.
\newblock In \emph{ICCV}, pages 1905--1914, 2021.

\bibitem[Wei et~al.(2025)Wei, Liu, Yuan, and Zhang]{wei2025perceive}
Hongyang Wei, Shuaizheng Liu, Chun Yuan, and Lei Zhang.
\newblock Perceive, understand and restore: Real-world image super-resolution with autoregressive multimodal generative models.
\newblock \emph{arXiv preprint arXiv:2503.11073}, 2025.

\bibitem[Wu et~al.(2023)Wu, Zhang, Liao, Chen, Hou, Wang, Sun, Yan, and Lin]{wu2023exploring}
Haoning Wu, Erli Zhang, Liang Liao, Chaofeng Chen, Jingwen Hou, Annan Wang, Wenxiu Sun, Qiong Yan, and Weisi Lin.
\newblock Exploring video quality assessment on user generated contents from aesthetic and technical perspectives.
\newblock In \emph{ICCV}, pages 20144--20154, 2023.

\bibitem[Wu et~al.(2024{\natexlab{a}})Wu, Sun, Ma, and Zhang]{wu2024one}
Rongyuan Wu, Lingchen Sun, Zhiyuan Ma, and Lei Zhang.
\newblock One-step effective diffusion network for real-world image super-resolution.
\newblock \emph{NeurIPS}, 37:\penalty0 92529--92553, 2024{\natexlab{a}}.

\bibitem[Wu et~al.(2024{\natexlab{b}})Wu, Yang, Sun, Zhang, Li, and Zhang]{wu2024seesr}
Rongyuan Wu, Tao Yang, Lingchen Sun, Zhengqiang Zhang, Shuai Li, and Lei Zhang.
\newblock Seesr: Towards semantics-aware real-world image super-resolution.
\newblock In \emph{CVPR}, pages 25456--25467, 2024{\natexlab{b}}.

\bibitem[Wu et~al.(2022)Wu, Wang, Li, and Shan]{wu2022animesr}
Yanze Wu, Xintao Wang, Gen Li, and Ying Shan.
\newblock Animesr: Learning real-world super-resolution models for animation videos.
\newblock \emph{NeurIPS}, 35:\penalty0 11241--11252, 2022.

\bibitem[Xie et~al.(2023)Xie, Wang, Shi, Gu, Dong, and Shan]{xie2023mitigating}
Liangbin Xie, Xintao Wang, Shuwei Shi, Jinjin Gu, Chao Dong, and Ying Shan.
\newblock Mitigating artifacts in real-world video super-resolution models.
\newblock In \emph{AAAI}, pages 2956--2964, 2023.

\bibitem[Xie et~al.(2025)Xie, Liu, Zhou, Zhao, Zhou, Zhang, Zhang, Yang, Yang, and Tai]{xie2025star}
Rui Xie, Yinhong Liu, Penghao Zhou, Chen Zhao, Jun Zhou, Kai Zhang, Zhenyu Zhang, Jian Yang, Zhenheng Yang, and Ying Tai.
\newblock Star: Spatial-temporal augmentation with text-to-video models for real-world video super-resolution.
\newblock \emph{arXiv preprint arXiv:2501.02976}, 2025.

\bibitem[Xue et~al.(2019)Xue, Chen, Wu, Wei, and Freeman]{xue2019video}
Tianfan Xue, Baian Chen, Jiajun Wu, Donglai Wei, and William~T Freeman.
\newblock Video enhancement with task-oriented flow.
\newblock \emph{IJCV}, 127\penalty0 (8):\penalty0 1106--1125, 2019.

\bibitem[Yang et~al.(2024{\natexlab{a}})Yang, Wu, Ren, Xie, and Zhang]{yang2024pixel}
Tao Yang, Rongyuan Wu, Peiran Ren, Xuansong Xie, and Lei Zhang.
\newblock Pixel-aware stable diffusion for realistic image super-resolution and personalized stylization.
\newblock In \emph{ECCV}, pages 74--91. Springer, 2024{\natexlab{a}}.

\bibitem[Yang et~al.(2021)Yang, Xiang, Zeng, and Zhang]{yang2021real}
Xi Yang, Wangmeng Xiang, Hui Zeng, and Lei Zhang.
\newblock Real-world video super-resolution: A benchmark dataset and a decomposition based learning scheme.
\newblock In \emph{ICCV}, pages 4781--4790, 2021.

\bibitem[Yang et~al.(2024{\natexlab{b}})Yang, He, Ma, and Zhang]{yang2024motion}
Xi Yang, Chenhang He, Jianqi Ma, and Lei Zhang.
\newblock Motion-guided latent diffusion for temporally consistent real-world video super-resolution.
\newblock In \emph{ECCV}, pages 224--242. Springer, 2024{\natexlab{b}}.

\bibitem[Zhang et~al.(2015)Zhang, Zhang, and Bovik]{zhang2015feature}
Lin Zhang, Lei Zhang, and Alan~C Bovik.
\newblock A feature-enriched completely blind image quality evaluator.
\newblock \emph{IEEE TIP}, 24\penalty0 (8):\penalty0 2579--2591, 2015.

\bibitem[Zhang and Yao(2024)]{zhang2024realviformer}
Yuehan Zhang and Angela Yao.
\newblock Realviformer: Investigating attention for real-world video super-resolution.
\newblock In \emph{ECCV}, pages 412--428. Springer, 2024.

\bibitem[Zhou et~al.(2024)Zhou, Yang, Wang, Luo, and Loy]{zhou2024upscale}
Shangchen Zhou, Peiqing Yang, Jianyi Wang, Yihang Luo, and Chen~Change Loy.
\newblock Upscale-a-video: Temporal-consistent diffusion model for real-world video super-resolution.
\newblock In \emph{CVPR}, pages 2535--2545, 2024.

\bibitem[Zhu et~al.(2025)Zhu, Jiang, Zhu, Zhang, Bull, and Zeng]{zhu2025blind}
Qiang Zhu, Yuxuan Jiang, Shuyuan Zhu, Fan Zhang, David Bull, and Bing Zeng.
\newblock Blind video super-resolution based on implicit kernels.
\newblock \emph{arXiv preprint arXiv:2503.07856}, 2025.

\end{thebibliography}
}


\end{document}